\documentclass{article}

\usepackage[preprint]{neurips_2021}

\usepackage{multirow}

\usepackage[utf8]{inputenc} 
\usepackage[T1]{fontenc}    
\usepackage{hyperref}       
\usepackage{url}            
\usepackage{booktabs}       
\usepackage{amsfonts}       
\usepackage{nicefrac}       
\usepackage{microtype}      
\usepackage{xcolor}         
\usepackage{graphicx}
\title{Facial Expression Recognition based on Multi-head Cross Attention Network}

\author{%
  Jae-Yeop Jeong, Yeong-Gi Hong, Daun Kim, and Jin-Woo Jeong\\
  Department of Data Science\\
  Seoul National University of Science and Technology\\
  Seoul, Korea \\
  \texttt{\{jaey.jeong, yghong, daun, jinw.jeong\}@seoultech.ac.kr} \\
   \And
   Yuchul Jung \\
   Department of Computer Science\\
   Kumoh National Institute of Technology \\
   Gumi, Korea \\
   \texttt{jyc@kumoh.ac.kr} \\
}

\begin{document}

\maketitle

\begin{abstract}
  Facial expression in-the-wild is essential for various interactive computing domains. In this paper, we proposed an extended version of DAN model to address the VA estimation and facial expression challenges introduced in ABAW 2022. Our method produced preliminary results of 0.44 of mean CCC value for the VA estimation task, and 0.33 of the average F1 score for the expression classification task.
\end{abstract}

\section{Introduction}

Recognition of facial expression is essential for various interactive computing domains, such as human-computer/machine interaction, human-robot interaction, and human-AI interaction. Previous studies on facial expression mainly utilized a set of human faces captured in a controlled setting, resulting in various limitations of the application in-the-wild. Recently, various works focusing on the affective behavior analysis in-the-wild have been introduced to realize the generation of trust, understanding and closeness between humans and machines in real life environments [17]. 

The 3rd competition on Affective Behavior Analysis in-the-wild (ABAW),  held in conjunction with the IEEE International Conference on Computer Vision and Pattern Recognition (CVPR) 2022, is a place where researchers can present their own contributions on the automatic analysis of human behavior and emotion recognition which is robust to video recording conditions, diversity of contexts, and timing of display [17]. The 3rd ABAW competition is based on the Aff-Wild2 database [2-8] 
which is an extension of the Aff-wild database [9] 
and consists of the following four tracks: 1) Valence-Arousal (VA) estimation, 2) Expression classification, 3) Action Unit (AU) detection, and 4) Multi-Task-Learning (MTL). In this paper, we describe our methods on VA estimation challenge and expression classification challenge and provide preliminary results. For the VA estimation challenge, totally 564 videos of around 2.8M frames that contain annotations in terms of valence and arousal (values ranged continousouly in [-1, +1]) were used. Similarly, for the expression classificaiton challenge, totally 548 videos of around 2.7M frames that contain annotations in terms of the 6 basic expressions (i.e., Anger, Disgust, Fear, Happy, Sad, Surprised), plus the neutral state, plus a category 'other' that denotes expressions/affective states other than the 6 basic ones were used. As evaluation metrics, mean Concordance Correlation Coefficient (CCC) of valence and arousal and the average F1 Score across all 8 categories were used for the VA estimation challenge and the expression classification challenge, respectively.
In this paper, we propose an extended version of DAN model based on ResNet with attention mechanisms proposed by [10] to solve the challenges mentioned above  
, and present the preliminary results on the official validation set. More precise and detailed results can be updated and added through the subsequent submissions to the competition.

\section{Method}

\subsection{Data Pre-processing}
The amount and the diversity of data is one of the most important key points for successful deep learning applications in terms of a model performance. However, as shown in Table 1, the Aff-wild2 dataset has a class imbalance problem, resulting in some emotional categories have far fewer images than others. To address this issue, we took two strategies, adding external databases and applying data augmentation. First, we used external facial expression databases, such as AffectNet [13], ExpW [14], and Ai-Hub vision dataset [15]. The sample images from each dataset can be found from Figure 2. As shown in the figure, Aff-wild2 and AffectNet share the same facial expression categories while ExpW and Ai-Hub datasets had only part of expression categories. The aforementioned databases were generally collected under in-the-wild setting. Note that Ai-hub dataset [15] is comprised of facial expressions taken by Korean actors in-the-wild. Among the classes included in the Ai-hub dataset, we only used a set of images with neutral, anger, fear, and surprise expressions.
Second, we employed various data augmentation techniques; color jitter, random crop, horizontal flip, color jitter with random crop, random crop with flip to prevent over-fitting. Finally, we cropped the face region of each image using DeepFace face detector [16] algorithm and then resized every patch into 224 x 224 scale. Table 1 shows the statistics of dataset we used when training the model.

\subsection{Model Architecture}
The overall architecture of our method is illustrated in Figure 1. Our method is based on the DAN approach which consists of the following two modules: a feature extractor and an attention phase. First, the feature extractor module extracts the intermediate visual features from input images with a discriminative loss function, named affinity loss, to maximize the classes margin. For feature extraction, we utilized a ResNet-50 network pretrained on VGGFace2 dataset.
Afterwards, a multi-head attention network consisting of a combination of a spatial attention unit and a channel attention unit takes the features and outputs an attention map. Finally, attention fusion network merges attention maps to be learned in an orchestrated fashion [10].
Second, after applying the process mentioned above, the final feature information is fed to fully connected layer and batch normalization layer. At the final step of the model architecture, we adopted a softmax layer with focal loss for expression classification task, and a tanh function with CCC loss for VA estimation task.
Finally, in this work, we applied a soft-voting based bagging approach (i.e., multiple variants models are trained and used for validation together).  

\begin{figure}[t!]
    \centering
    \includegraphics[width=\columnwidth]{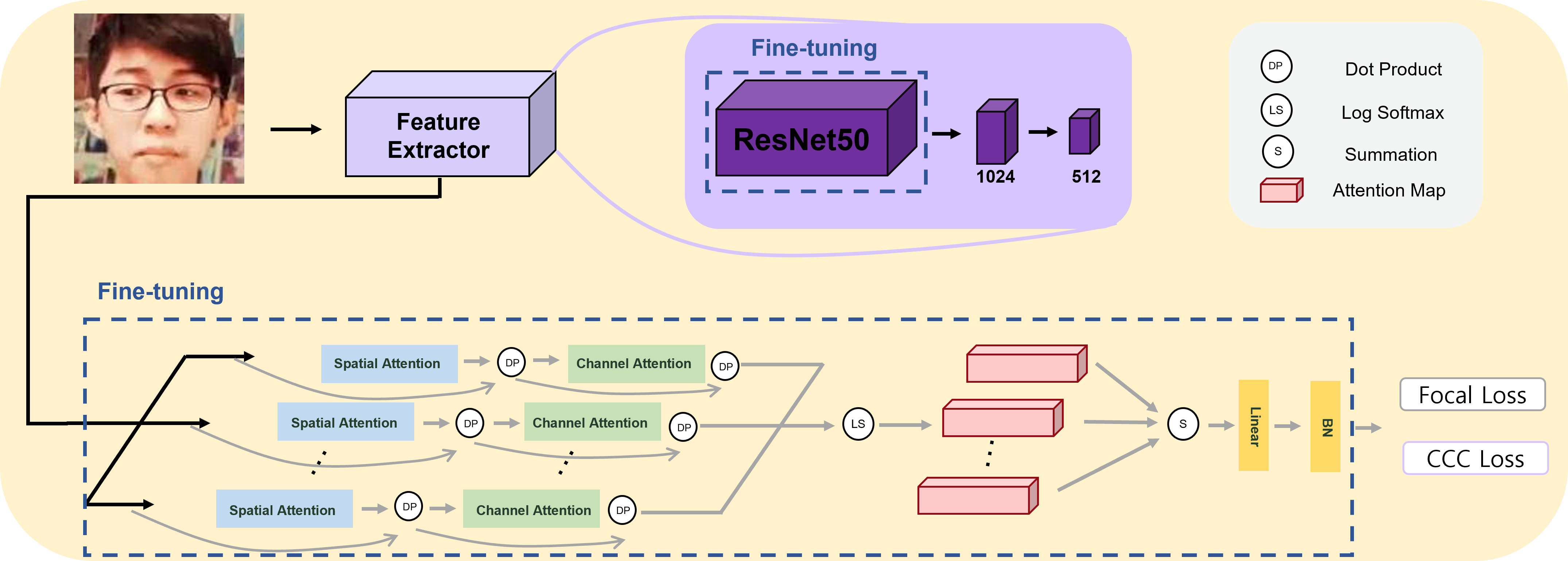}
    \caption{Overview of the architecture used in this study}
    \label{fig:architecture}
\end{figure}

\begin{table}[t!]
\caption{Data statistics}
\begin{tabular}{l|llllllll}
\hline
\multicolumn{1}{c|}{Database} & \multicolumn{1}{c}{Neutral} & \multicolumn{1}{c}{Anger} & \multicolumn{1}{c}{Disgust} & \multicolumn{1}{c}{Fear} & \multicolumn{1}{c}{Happy} & \multicolumn{1}{c}{Sad} & \multicolumn{1}{c}{Suprise} & \multicolumn{1}{c}{Other} \\ \hline
Aff-Wild2                      & 177,498                     & 16,573                    & 10,810                      & 9,080                    & 95,633                    & 79,862                  & 31,637                      & 165,866                    \\
AffectNet                      & 74,874                      & 24,882                    & 3,803                       & 6,378                    & 134,415                   & 25,459                  & 14,090                      & 3,750                      \\
ExpW                           & 34,883                      & 3,671                     & 3,395                       & 1,088                    & 30,537                    & 10,559                  & 7,060                       & 0                          \\
AI-Hub                           &                       & 59,696                     &                        & 59,262                    &                     &                   & 59,643                       &                           \\\hline
\end{tabular}
\label{data_stat}
\end{table}

\section{Results}
All the experiments were conducted using a GPU server with six NVIDIA RTX 3090 GPUs, 128 GB RAM, and an Intel i9-10940X CPU. We used Pytorch framework for the implementation/modification, training and evaluation of the model. Our preliminary results on the official validation set for the VA estimation task was 0.44, and 0.33 for the expression classification task. 

\begin{figure}[t]
    \centering
    \includegraphics[width=\columnwidth]{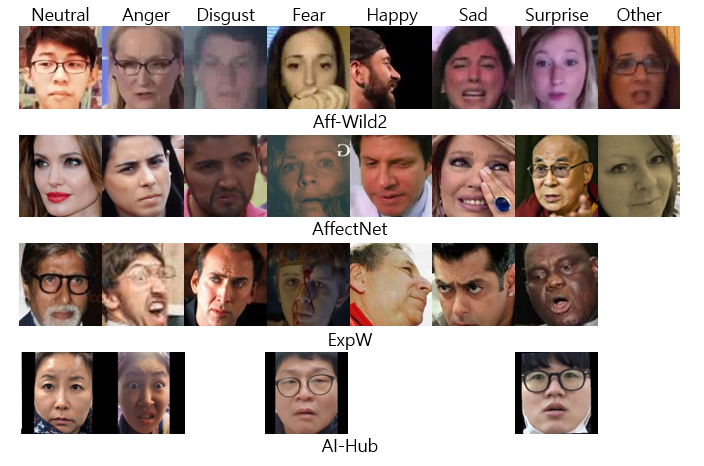}
    \caption{Datasets used in our study}
  
\end{figure}

\begin{table}[t]
\centering
\caption{Hyper-parameter setting}
\begin{tabular}{ll}
\hline
Hyper-parameter        & Value   \\ \hline
Learning Rate          & $1e^{-4}$ \\
Batch Size             & 1024    \\
Epochs                 & 8       \\
Optimizer Weight Decay & $1e^{-4}$ \\
Number of Head         & 4       \\ \hline
\end{tabular}

\end{table}

\section{Conclusion}

In this paper, we proposed an extended version of DAN model to address the VA estimation and facial expression challenges introduced in ABAW 2022. Our method produced preliminary results of 0.44 of mean CCC value for the VA estimation task, and 0.33 of the average F1 score for the expression classification task. The details and results may be updated after submission of this paper to arxiv.

\section*{References}

{
[1] Kollias, Dimitrios : {\it ABAW: Valence-Arousal Estimation, Expression Recognition, Action Unit Detection \& Multi-Task Learning Challenges.} arXiv:2202.10659 , 2021

[2] Kollias, Dimitrios and Zafeiriou, Stefanos : {\it Analysing Affective Behavior in the second ABAW2 Competition.}\ pp.\ 3652--3660\ ICCV, 2021

[3] Kollias, D and Schulc, A and Hajiyev, E and Zafeiriou, S: {\it Analysing Affective Behavior in the First ABAW 2020 Competition.}\ pp.\ 794--800 IEEE FG, 2020

[4] Kollias, Dimitrios and Sharmanska, Viktoriia and Zafeiriou, Stefanos: {\it Distribution Matching for Heterogeneous Multi-Task Learning: a Large-scale Face Study.} arXiv:2105.03790, 2021

[5] Kollias, Dimitrios and Zafeiriou, Stefanos: {\it Affect Analysis in-the-wild: Valence-Arousal, Expressions, Action Units and a Unified Framework.} arXiv:2105.03790, 2021

[6] Kollias, Dimitrios and Zafeiriou, Stefanos: {\it Expression, Affect, Action Unit Recognition: Aff-Wild2, Multi-Task Learning and ArcFace.} BMVC, 2019

[7] Kollias, Dimitrios and Sharmanska, Viktoriia and Zafeiriou, Stefanos: {\it Face Behavior a la carte: Expressions, Affect and Action Units in a Single Network.} arXiv:1910.04855, 2019

[8] Kollias, Dimitrios and Tzirakis, Panagiotis and Nicolaou, Mihalis A and Papaioannou, Athanasios and Zhao, Guoying and Schuller, Bj{\"o}rn and Kotsia, Irene and Zafeiriou, Stefanos: {\it Deep Affect Prediction in-the-wild: Aff-Wild Database and Challenge, Deep Architectures, and Beyond.}\ pp.\ 1--23\ International Journal of Computer Vision (IJCV), 2019

[9] Zafeiriou, Stefanos and Kollias, Dimitrios and Nicolaou, Mihalis A and Papaioannou, Athanasios and Zhao, Guoying and Kotsia, Irene : {\it Aff-Wild: Valence and Arousal in-the-wild Challenge.}\ pp.\ 1980--1987\ IEEE, 2017

[10] Zhengyao Wen, Wenzhong Lin, Tao Wan*, Ge Xu in Minjiang University, China : {\it Distract Your Attention: Multi-head Cross Attention Network for Facial Expression Recognition.} arXiv:2109.07270 ,2019 

[11] Kaiming He, Xiangyu Zhang, Shaoqing Ren, Jian Sun in Microsoft Research : {\it Deep Residual Learning for Image Recognition.} arXiv:1512.03385 , 2015

[12] Tsung-Yi Lin, Priya Goyal, Ross Girshick, Kaiming He, Piotr Dollár in Facebook AI Research (FAIR) : {\it Focal Loss for Dense Object Detection.} arXiv:1708.02002 , 2018
 
[13] Ali Mollahosseini, Behzad Hasani, Mohammad H. Mahoor : {\it AffectNet: A Database for Facial Expression, Valence, and Arousal Computing in the Wild.} IEEE Transactions on Affective Computing , 2017

[14] Zhanpeng Zhang, Ping Luo, Chen Change Loy, Xiaoou Tang in Multimedia Laboratory, Department of Information Engineering, The Chinese University of Hong Kong : {\it Learning Social Relation Traits from Face Images} ICCV , 2015

[15] AI-HUB Dataset, https://aihub.or.kr/aidata/27716

[16] Yaniv Taigman, Ming Yang, Marc’Aurelio Ranzato in Facebook AI Research , \ Lior Wolf in Tel Aviv University : {\it DeepFace: Closing the Gap to Human-Level Performance in Face Verification.} IEEE, 2014

[17] Dimitrios Kollias : {\it ABAW: Valence-Arousal Estimation, Expression Recognition, Action Unit Detection \& Multi-Task Learning Challenges. } arXiv:2202.10659, 2022
}

\end{document}